\documentclass[sigconf]{acmart}

\usepackage{booktabs} % For formal tables

\setcopyright{rightsretained}
\acmConference[GECCO '21]{the Genetic and Evolutionary Computation Conference 2021}{July 10--14, 2021}{Lille, France}
\acmYear{2021}
\copyrightyear{2021}

\acmPrice{15.00}

\usepackage[ruled,vlined]{algorithm2e}
\usepackage{subcaption}

\newcommand{\figref}[1]{Fig.~\ref{fig:#1}}
\newcommand{\tabref}[1]{Table~\ref{tab:#1}}
\newcommand{\algref}[1]{Algorithm~\ref{alg:#1}}
\newcommand{\ie}{\emph{i.e.}}

\newcommand{\etc}{\emph{etc.}}

\copyrightyear{2021}
\acmYear{2021}
\setcopyright{acmcopyright}\acmConference[GECCO '21 Companion]{2021 Genetic and Evolutionary Computation Conference Companion}{July 10--14, 2021}{Lille, France}
\acmBooktitle{2021 Genetic and Evolutionary Computation Conference Companion (GECCO '21 Companion), July 10--14, 2021, Lille, France}
\acmPrice{15.00}
\acmDOI{10.1145/3449726.3463189}
\acmISBN{978-1-4503-8351-6/21/07}

\begin{document}
\title{Evolving Neural Selection with Adaptive Regularization}
% \titlenote{Produces the permission block, and
% \subtitle{Subtitle}
% \subtitlenote{The full version of the author's guide is available as
%   \texttt{acmart.pdf} document}

%%% The submitted version for review should be ANONYMOUS
\author{Li Ding}
% \authornote{Dr.~Trovato insisted his name be first.}
\affiliation{%
  \institution{University of Massachusetts Amherst}
  \institution{Massachusetts Institute of Technology}
  % \streetaddress{P.O. Box 1212}
}
\email{liding@umass.edu}

\author{Lee Spector}
% \authornote{Dr.~Trovato insisted his name be first.}
\affiliation{%
  \institution{Amherst College}
  \institution{Hampshire College}
  \institution{University of Massachusetts Amherst}
  % \streetaddress{P.O. Box 1212}
}
\email{lspector@amherst.edu}

% \authornote{The secretary disavows any knowledge of this author's actions.}
% \affiliation{%

% \authornote{This author is the
%   one who did all the really hard work.}
% \affiliation{%

% \affiliation{%
% \affiliation{%
% \affiliation{%
%   \country{China}
% }

% \affiliation{%


% \author{Julius P.~Kumquat}

% % The default list of authors is too long for headers.
\renewcommand{\shortauthors}{L. Ding and L. Spector}

\begin{abstract}
  Over-parameterization is one of the inherent characteristics of modern deep
  neural networks, which can often be overcome by leveraging regularization
  methods, such as Dropout~\cite{srivastava2014dropout}. Usually, these methods
  are applied globally and all the input cases are treated equally. However,
  given the natural variation of the input space for real-world tasks such as
  image recognition and natural language understanding, it is unlikely that a
  fixed regularization pattern will have the same effectiveness for all the
  input cases. In this work, we demonstrate a method in which the selection of
  neurons in deep neural networks evolves, adapting to the difficulty of
  prediction. We propose the Adaptive Neural Selection (ANS) framework, which
  evolves to weigh neurons in a layer to form network variants that are suitable
  to handle different input cases. Experimental results show that the proposed
  method can significantly improve the performance of commonly-used neural
  network architectures on standard image recognition benchmarks. Ablation
  studies also validate the effectiveness and contribution of each component in
  the proposed framework.
\end{abstract}

  


\maketitle

\section{Introduction}

Modern neural networks usually adopt deep architectures with millions of
parameters and connections. One common issue with such designs is
over-parameterization~\cite{nakkiran2019deep}, in which case the number of
parameters is too large for the task. This situation often results in
overfitting, which hurts the generalization capability of the model at test
time, as the deep network tends to learn specific representations and memorize
all the training samples. A general strategy to resolve this problem is to use
regularization methods that penalize the model using too many parameters. For
neural networks specifically, Dropout~\cite{srivastava2014dropout} has been a
common method which randomly selects a certain ratio of neurons from a layer of
the network and temporarily removes them along with all their incoming and
outgoing connections. For every training iteration the network will use fewer
parameters in a probabilistic fashion, and thus can be prevented from learning
specific representations for each training samples. 

There are a number of studies proposing variants of Dropout that impose certain
structures to specific models or tasks. However, most of them are highly
specialized and can not be easily adapted to different network architectures or
training protocols. In the image recognition domain, the most common way of
using Dropout is to apply it to the fully-connected layers or simply the last
fully-connected layer before the classification layer. Some recent studies also
propose variants that perform Dropout on convolutional layers by dropping
neurons in particular structures. For example, DropPath~\cite{zoph2018learning}
forms multi-branched convolutional cells and drops one random branch each time.
DropBlock~\cite{ghiasi2018dropblock} selects contiguous squares of neurons in
the convolutional layers and drops them. However, these methods usually require
specific tuning of the architecture and parameters based on the model and
dataset. In other words, the same dropout pattern does not work as well on a
different model or dataset. More recently,
AutoDropout~\cite{pham2021autodropout} proposes to automate the process of
designing dropout patterns in a reinforcement learning setting. While it shows
promising improvement over numerous benchmarks, the cost of searching dropout
patterns can be over a magnitude higher than the training job itself, making the
method less practical in real-world tasks.

\begin{figure*}
  \begin{subfigure}{.48\textwidth}
    \centering
    \includegraphics[width=.85\linewidth]{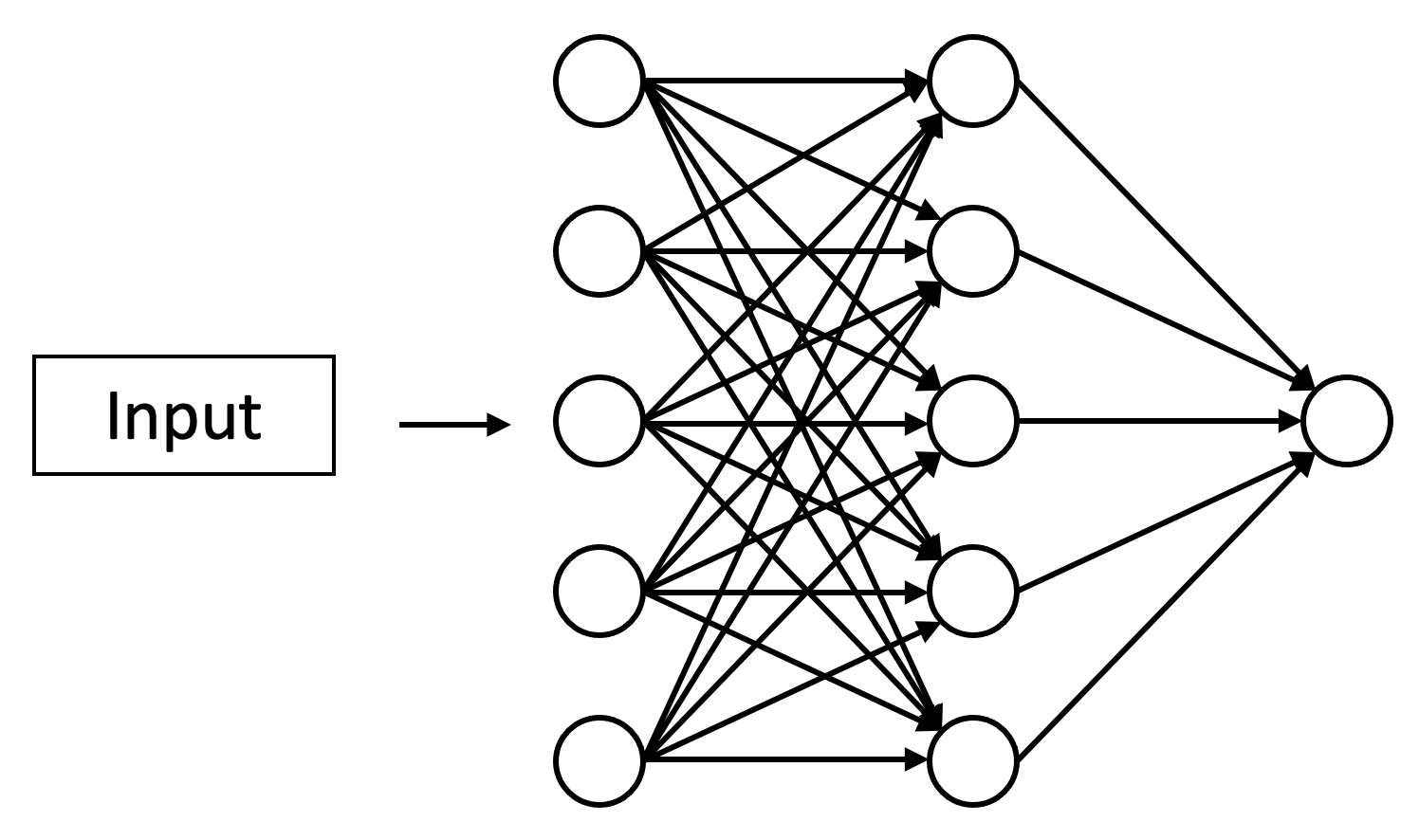}  
    \caption{Ordinary Neural Network}
    \label{fig:nn}
  \end{subfigure}
  \begin{subfigure}{.48\textwidth}
    \centering
    \includegraphics[width=.85\linewidth]{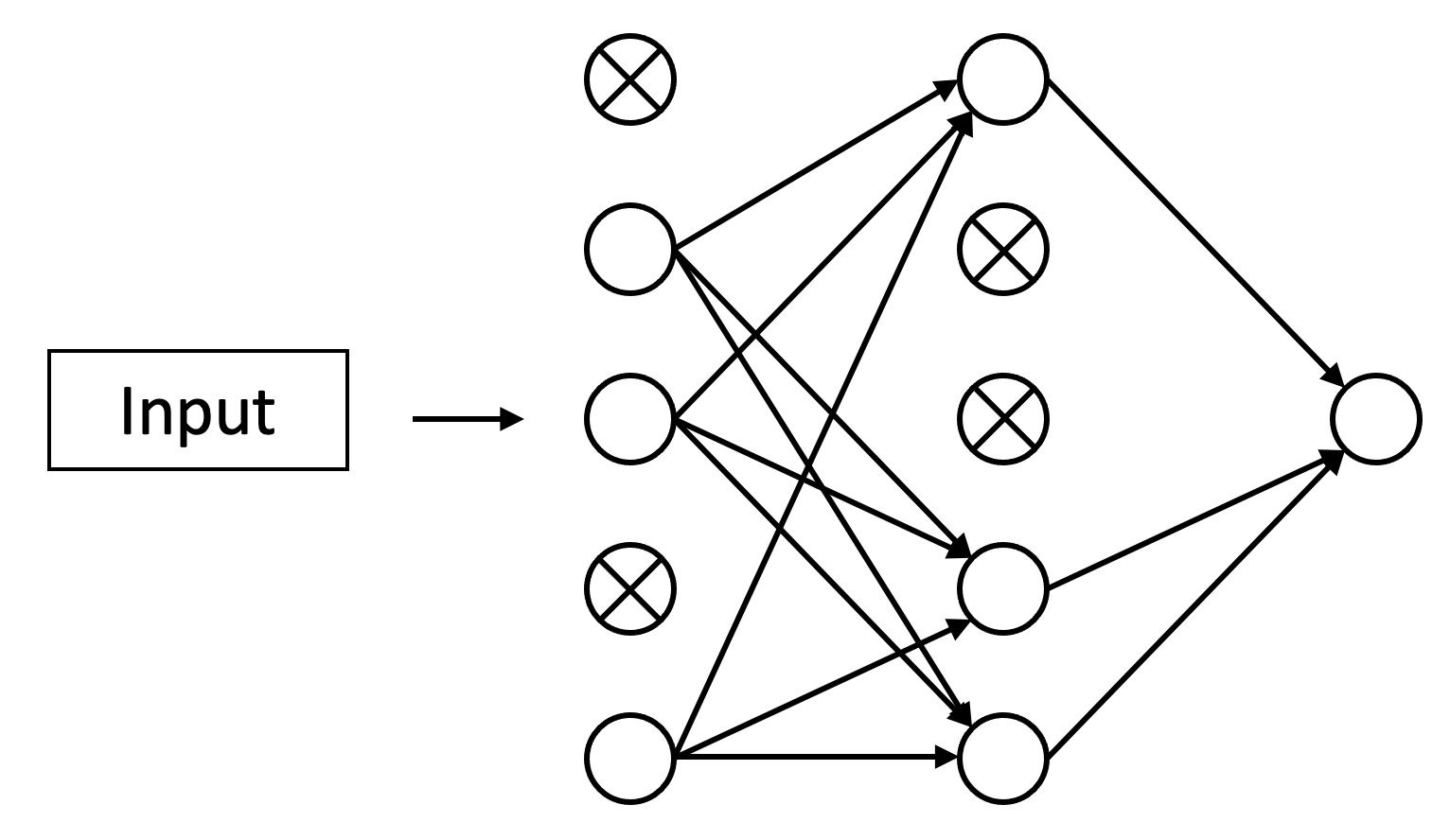}  
    \caption{Neural Network with Dropout~\cite{srivastava2014dropout}}
    \label{fig:drop}
  \end{subfigure}
  \\
  \vspace{4em}
  \begin{subfigure}{\textwidth}
    \centering
    \includegraphics[width=.93\linewidth]{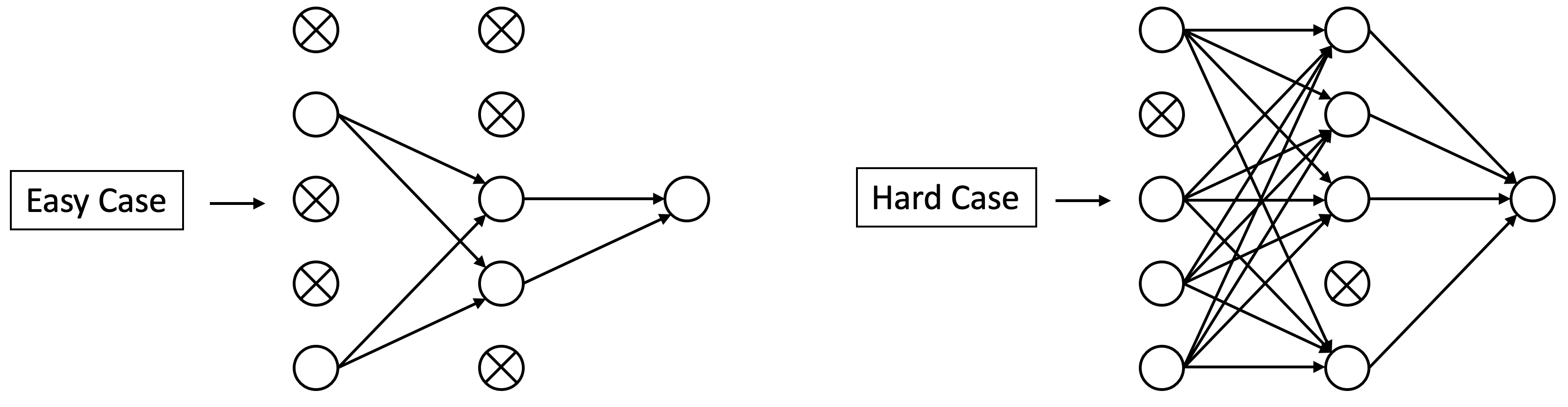}  
    \caption{Neural Network with Adaptive Neural Selection (ANS)}
    \label{fig:ans}
  \end{subfigure}
  \caption{Illustration of Adaptive Neural Selection (ANS) framework. a) An
  ordinary neural network with fully connected layers. b) A neural network after
  applying dropout: The neurons and connection are dropped at a random ratio for
  all the input during training. c) A neural network with the proposed ANS
  framework: The neurons are adaptively weighted and the weights are predicted
  by a self-attention module regarding the input. For easier cases, more neurons
  are dropped to prevent overfitting.}
  \label{fig:overview}
\end{figure*}

In general, the main idea of these previous works is, while neural networks
usually show better performance when going deeper and having more parameters
(assuming that there exists a sufficiently large amount of data), they are not
necessarily required to use all the neurons to make good predictions, \ie, the
activation units in a layer can be sparse. Such sparsity in practice has a
regularization effect on over-parameterization. However, most previous works
tend to explore specific ways to apply regularization, and use the same
regularization pattern over all the input training samples, which may not be
optimal because the complexity and variation of input samples are essentially
different. For example, distinguishing images of coarse categories, such as dogs
vs. cats, is much easier than distinguishing more fine-grained categories, such
as two certain species of dogs, because the latter requires learning a more
complex combination of visual representations. Thus, another potential way to
improve regularization in deep neural networks is to properly allocate the right
number of neurons based on each input case, and such a number is just enough for
the model to make a correct prediction.

In this work, we aim to explore this regime of keeping essential neurons and
dropping out unnecessary ones based on the input data, in order to form simple
and powerful representations that generalize well to unseen test cases. An
alternative way to view this goal is to generate different network variants that
are selected to handle each input sample accordingly, because the samples are
likely to require different numbers of neurons to form reasonable
representations. Inadequate resources may cause undermined model performance,
and on the other hand, overwhelming resources are likely to make the model
overfit. As similar problems are well studied in the literature of evolutionary
computing~\cite{eiben:2003:book}, we consider a novel strategy that evolves
selections of neurons to form neural network variants in parallel with the
training process of the neural network. The fitness of each network variant is
measured by the model performance and the number of neurons used for the
prediction. We term this framework Adaptive Neural Selection (ANS), which
consists of two parts: a self-attention module and an adaptive regularization
mechanism. 

The self-attention module is intended to perform selection of neurons by
learning a variable weight to be imposed on each neuron. The attention mechanism
has been shown in several previous studies~\cite{vaswani2017attention,
dai2019transformer, hu2018squeeze} to help deep networks focus on essential
features and neurons. While the self-attention module can be trained end-to-end
in parallel with the neural network through gradient descent, it does not have
an objective to adapt neural selection based on the difficulty of predicting
each input sample. Thus, we further introduce an adaptive regularization method.
This regularization works as a selection pressure that penalizes the
self-attention module if it is selecting more neurons than required. The
penalization is adaptive to different training samples in regard to the
difficulty of performing the task. For example, given a batch of images, if the
classification accuracy is low, that means the model has not fully learned good
representations of these cases, and thus it is allowed to use more neurons to
explore a more complex feature space. The two mechanisms work together to form
the Adaptive Neural Selection (ANS) framework, which evolves different
sub-network variants by learning the adaptive selection of neurons. ANS can be
used on any neural network architecture to adapt its complexity and structure to
handle the variety of input cases. 

We implement the proposed ANS framework and test it for the image classification
task, which is one of the most common tasks for modern deep neural networks. The
experiments show that ANS significantly improves commonly-used deep ConvNet
architectures on standard benchmarks. On CIFAR-100, ANS improves the test
accuracy of ResNet-50 from $76.65\%$ to $78.08\%$, and VGG-16 from $62.30\%$ to
$65.21\%$, outperforming Dropout~\cite{srivastava2014dropout} by a significant
margin. Ablation studies also validate that the two components both make a good
contribution to the improvement. 

In the following sections, we first summarize the related work in the area of
machine learning and evolutionary computing. Then we describe the ANS framework
and its components. This is followed by experimental results showing the
performance of our method. Finally, we conclude by summarizing the results and
listing directions for future work.

\section{Related Work}

Within the broader context of machine learning research, this paper is in line
with existing works on regularization methods for neural
networks~\cite{yun2019cutmix, devries2017improved, gal2016dropout,
huang2016deep, nakkiran2019deep, srivastava2014dropout, ghiasi2018dropblock,
pham2021autodropout, zoph2018learning}. Most of the recent methods such as
Dropout~\cite{srivastava2014dropout} and its variants require either specific
and careful tuning of the structure based on the datasets or
tasks~\cite{ghiasi2018dropblock, zoph2018learning}, or expensive computation
cost for searching good patterns of regularization~\cite{pham2021autodropout}.
Instead of expanding on the current Dropout mechanism, our work explores the
problem of regularizing over-parameterization in neural networks from a
evolutionary computing perspective. In addition, our work is also related to
automated data augmentation~\cite{zoph2020learning, lim2019fast,
xie2019unsupervised,cubuk2018autoaugment, li2020feature}. Unlike most data
augmentation methods that are usually applied specifically to the type of input
data, our method works on the high-level representations of neural networks. It
employs a more general design philosophy, and thus the same mechanism can be
easily applied to different data types and tasks. 

Our work is inspired by the parent selection methods~\cite{eiben:2003:book,
koza:book, helmuth2014solving,LaCava:EC,aenugu2019lexicase,McKay:2000:GECCO} in
evolutionary computing. More specifically, our method is related to the Implicit
Fitness Sharing~\cite{McKay:2000:GECCO} and Lexicase
selection~\cite{helmuth2014solving,LaCava:EC,aenugu2019lexicase}, in which hard
cases are either weighted more heavily in aggregate fitness measures, or
required to be correctly predicted by some model variants in every generation,
when the training samples are ordered in random sequences. The proposed ANS
framework is designed to put selection pressure to guide the neural networks to
use the right amount of neurons depending on model performance. Since the neural
networks usually require batched training samples to work with large-scale
dataset, we follow the general idea in \cite{aenugu2019lexicase} to make the
selection framework suitable to be trained with batched samples. 

We also take advantage of the attention mechanism, which has been proved to be
useful in improving neural network performance with different architectural
designs~\cite{vaswani2017attention, dai2019transformer, hu2018squeeze} in the
context of deep learning. However, most of these methods rely on gradient-based
training of the attention layers, in which case the updates of weights in the
attention layers are correlated with the updates of weights in the neural
network. This will result in co-adaption of weights in both the attention layers
and other layers in the neural network, which may still cause overfitting since
the network can still learn to memorize all the training samples. On the other
hand, our work evolves the self-attention module based on a combination of the
proposed adaptive regularization method and gradients from the model prediction
loss. In this way, we can prevent the co-adaption of attention and neural
network weights, and thus have a stronger regularization effect on the network.

Our work is also closely-connected to prior research on neural
evolution~\cite{stanley2002evolving, real2017large, real2019regularized,
stanley2009hypercube,miikkulainen2019evolving}, and neural architecture
search~\cite{liu2018darts,
pham2018efficient,liu2018progressive,cai2019automl,liu2017hierarchical}. The
proposed ANS framework assumes a fixed neural network architecture and evolves
different model variants to handle different input cases, which is different
from the general neural evolution and architecture search methods that aim to
evolve the network architectures. Our method can be further combined with neural
evolution methods to co-evolve network variants and network architectures at the
same time.

\section{Methods}

In this section, we introduce the Adaptive Neural Selection (ANS) framework. ANS
aims to evolve the behavior of selecting certain neurons from the deep neural
networks to perform the prediction task based on the input training samples. The
framework consists of two parts: a self-attention module and an adaptive
regularization mechanism. An illustration of our method comparing to ordinary
neural networks and the Dropout~\cite{srivastava2014dropout} is shown in
\figref{overview}.

\subsection{Ordinary Neural Network}

Consider a neural network with $c$ fully connected layers. Let $l \in
\{1,\cdots, c\}$ denote the index of hidden layers. Let $x^{(l)}$ denote the
input vector to layer $l$, and $y^{(l)}$ denote the output vector from layer $l$
before activation. $W^{(l)}$ and $b^{(l)}$ are the weight matrix and bias vector
at layer $l$. The forward pass operation of an ordinary neural network at layer
$l$ can be described as 

\begin{align}
  y^{(l)} &= W^{(l)}x^{(l)}+b^{(l)}\\
  x^{(l+1)} &= f(y^{(l)})
\end{align}
where $f(\cdot)$ is the activation function, such as the ReLU function $f(x) = \max(0,x)$.

The neural network is usually trained using back-propagation and gradient
descent. We use $L_{grad}$ to denote the common gradient-based loss functions.
For example, the image classification task usually use cross-entropy loss on the
one-hot encoded output, described as
\begin{align}
  L_{grad} &= -\sum_{Y\in \mathbb{Y}}(Y\cdot\log\hat{Y})\\
  \hat{Y} &= f(y^{(c)})
\end{align}
where $Y$ is the one-hot encoded ground truth label in the set $\mathbb{Y}$ of
all the training labels, and $\hat{Y}$ is the corresponding network's prediction.
The prediction is usually calculated using the softmax function as $f$ on the
last layer's output.

For neural networks that have specific layers for feature extraction, such as
convolutional neural networks (ConvNets), we also use the above formulation for
the fully-connected layers in the network, which usually comes after the feature
extraction layers.

\subsection{Self-attention Module}

To add the self-attention module, we use $z^{(l)}$ to denote an attention layer
with another pair of weight matrix $W_a^{(l)}$ and bias vector $b_a^{(l)}$. The
forward pass operation of a neural network with attention module can be
described as
\begin{align}
  y^{(l)} &= W^{(l)}x^{(l)}+b^{(l)}\\
  z^{(l)} &= W_a^{(l)}f(y^{(l)})+b_a^{(l)}\\
  x^{(l+1)} &= g(z^{(l)}) f(y^{(l)})
\end{align}
where $f(\cdot)$ is the activation function and $g(\cdot)$ is the gating
function on the attention weights. In this work, we use the soft attention
mechanism which uses the sigmoid function $g(x) = 1/(1+e^{-x})$. 

By adding the self-attention module, the neuron activations $f(y^{(l)})$ are now
regularized by the attention units $g(z^{(l)})$. Since we have $g(z^{(l)})\in
(0,1)$ for the sigmoid function, the neurons become selective according to the
weights comparing to the ordinary neural network. We can also represent the
ordinary neural network with this formulation by having $g(z^{(l)})=1$ for every
neuron, and Dropout can be viewed as having random binary values of
$g(z^{(l)})$. Thus, the self-attention module can achieve a larger space of
sub-network structures comparing to Dropout.

While the weight matrix $W_a^{(l)}$ and bias vector $b_a^{(l)}$ for the
attention layer can be trained in parallel through normal gradient descent and
back-propagation, the resulting selection of neurons may not work well as a
regularization for the neural network. As the self-attention module is a fully
connected layer extended from the neural network, it may co-adapt with the
neural network weight and thus result in overfitting. To solve this issue, we
further introduce an adaptive regularization mechanism for the self-attention
module, as describe in the next subsection.

\subsection{Adaptive Regularization for Neural Selection}

In the context of modern deep neural networks, with similar architectures and
training protocols, better performance is usually achieved by network variants
with deeper structures and parameters. However, the overfitting problem is also
more likely to occur especially when training large neural networks with
insufficient data. One thing that people usually do not know in advance and thus
require lots of experiments and tuning is how many parameters works the best on
the given dataset, that will not cause overfitting. A common way to resolve this
issue is to have regularization on the neural network, and penalize it for
having or relying on too many parameters. For example, Dropout forces the
network to only use a random number of neurons, \ie, a sub-network, for
prediction during training, and Weight Decay penalizes complexity by adding all
the parameters to the loss function. 

Most of these methods apply equally to all the training and testing samples,
assuming that all the input samples follow the same distribution, and the neural
network is trained to work with that distribution. However, some input samples
have higher complexity, and thus may need more neurons to form a good
representation. To explore this regime, we propose an adaptive regularization
mechanism. For layer $l$ in the neural network, we use $g(z^{(l)})$ to denote
the self-attention units. The regularization is calculated as
\begin{align}
  L_{reg} = \gamma \cdot \frac{1}{k} \lVert g(z^{(l)}) \rVert_1
\end{align}
where $\gamma$ is a variable that controls the extent of regularization and $k$
is the number of all the neurons as a normalization factor.

We also introduce a strategy to calculate $\gamma$ depending on the model
performance on current input samples. The idea is that for better training
performance on a batch of data, the chance of overfitting is higher, and we thus
need greater regularization to guide the network to use fewer neurons that may
form more generalizable representations. To achieve this, we calculate $\gamma$
as
\begin{align}\label{eq:gamma}
  \gamma = \alpha \cdot M^{\beta}
\end{align}
where $\alpha$ and $\beta$ are hyperparameters controlling the scale and
variance of $\gamma$. $M$ denotes a non-negative metric of model performance on
current input samples. In the context of image classification, for example, we
use batch accuracy as the metric where $M\in [0,1]$. With this formulation, we
have $\lambda$ becomes larger when the model performance becomes better, and
thus enforcing greater regularization on the neural networks. 

\subsection{Evolving Adaptive Neural Selection}

Here we introduce the Adaptive Neural Selection (ANS) framework that combines
the self-attention module and the adaptive regularization method. Inspired by
parent selection methods in genetic programming, the ANS framework introduces a
selection pressure to the current neural network by injecting regularization on
the number of neurons currently being used for the input training sample. By
putting the attention weights on a set of neurons for each training sample, the
selection actually evolves different sub-network variants adaptively, and such
evolution improves as the training process proceeds. The fitness of neuron
selection is evaluated by the current model performance, and the way we optimize
the neural selection is to select as few neurons as possible depending on the
difficulty of predicting the input sample, in order to prevent overfitting on
trivial cases or underfitting on hard ones. The full training procedure is
described in \algref{ans}. 

\begin{algorithm}[t]
  \SetAlgoLined
  \KwData{training samples}
  \textbf{Parameters:} batch size, $\alpha$, $\beta$, learning rate, number of training epoches\\
  \While{current epoch < number of training epoches}{
    samples := list of training samples in random order\;
    \While{samples is not empty}{
      batch-sample := a batch of training samples from all samples\;
      feed forward the neural network with batch-sample, get batch-predictions\;
      $L_{grad}$ := the cross-entropy loss calculated using the ground truth labels\;
      $M$ := the batch accuracy (number of correct predictions / batch size)\;
      $L_{reg} := \gamma \cdot \frac{1}{k} \lVert g(z^{(l)}) \rVert_1$ where $\gamma = \alpha \cdot M^{\beta}$\;
      Optimize the neural network with objective function $J := L_{grad}+L_{reg}$, do back-propagation with learning rate.
    }
    decay learning rate if applicable\;
  }
  \caption{Evolving Adaptive Neural Selection}
  \label{alg:ans}
\end{algorithm}

During the procedure of evolution, the framework trains the neural networks by
using a combination of error gradient and regularization. With the help of
gradient-based learning, the neural selection can be efficiently optimized and
quickly reach convergence. In addition, since most neuroevolutionary methods
tends to use non-gradient methods for optimization, such as Gaussian mutation,
our framework can be easily extended to apply those techniques as well, by using
the $L_{reg}$ term for fitness evaluation. However, in order to show the
effectiveness of the proposed framework, we choose to not use those non-gradient
optimization methods to ensure fair comparison to the ordinary gradient-trained
models.

\section{Experiments}

In this section, we implement and test the proposed ANS framework to common
ConvNet architectures for the supervised image classification task. We compare
our method against the Dropout method with different parameter
settings. We also perform ablation study on the effect of hyperparameters of
ANS.

\subsection{ConvNet Architectures}

We use the ResNet-50~\cite{he2016deep} and VGG-16~\cite{simonyan2014very} models
for the experiments. Both of the networks are commonly used in various image
recognition problems and have been extended to other tasks such as semantic
segmentation and object detection. While there are several recent works proposing
better performing architectures with novel techniques, we choose to use these two
common architectures to better illustrate the performance of our method, which
can be further extended to other methods as well.

\subsection{Datasets and Metrics}

We use the CIFAR-10 and CIFAR-100 datasets~\cite{krizhevsky2009learning} for the
experiments. The CIFAR-10 dataset consists of 60000 32x32 colour images in 10
classes, with 6000 images per class. There are 50000 training images and 10000
test images. CIFAR-100 has 100 classes containing 600 images each. Similarly,
there are 500 training images and 100 testing images per class. The classes are
mutually exclusive. Both datasets serve as standard benchmarks for image
recognition.

Since both datasets come with a pre-defined testing set, and the classes are
well-balanced, we adopt the mean prediction accuracy as the metric. 

\subsection{Experiment Configuration}

We use similar training configurations that are commonly adopted in prior
works~\cite{devries2017improved, he2016deep}. For all the experiments, we use a
batch size of $64$ and a total number of epoches of $120$. The optimization
method is stochastic gradient descent with Nesterov momentum of $0.9$, with
weight decay of $1e-4$. The initial learning rate is $0.1$ for ResNet-50, and
$0.01$ for VGG-16, given the fact that VGG-16 has more parameters and thus needs
a smaller learning rate to prevent divergence. We decrease the learning rate by
a factor of $0.1$ at epoch $60$ and $90$ for better convergence. All the models
are trained from scratch with standard initialization methods as in the original
work. We also follow the common data augmentation pipeline in \cite{he2016deep},
which is random cropping and horizontal flipping. All the training jobs are
implemented with PyTorch~\cite{paszke2019pytorch} library and performed on
cluster nodes with a single GPU. The test results are obtained on predicting
with the single image only, without ensembling.

\subsection{Image Classification Results}

\begin{table}
  \caption{Classification results with ResNet-50 on the test sets of CIFAR-10 and CIFAR-100.}
  \label{tab:resnet}
  \begin{tabular}{lrr}
    \toprule
    Method & CIFAR-10 & CIFAR-100\\
    \midrule
    Vanilla ResNet-50  &  0.9418  &   0.7665\\
    Dropout (p = 0.3)  &  0.9421  &   0.7587\\
    Dropout (p = 0.5)  &  0.9392  &   0.7611\\
    ANS (ours)         &  \textbf{0.9498}  &   \textbf{0.7808}\\
  \bottomrule
\end{tabular}
\end{table}

\begin{table}
  \caption{Classification results with VGG-16 on the test sets of CIFAR-10 and CIFAR-100.}
  \label{tab:vgg}
  \begin{tabular}{lrr}
    \toprule
    Method & CIFAR-10 & CIFAR-100\\
    \midrule
    Vanilla VGG-16     &  0.9046  &   0.6230\\
    Dropout (p = 0.3)  &  0.9071  &   0.6455\\
    Dropout (p = 0.5)  &  0.9070  &   0.6457\\
    ANS (ours)         &  \textbf{0.9109}  &   \textbf{0.6521}\\
  \bottomrule
\end{tabular}
\end{table}

We apply the ANS framework on ResNet-50 and VGG-16. For comparison, we also
implement the baseline method of the vanilla architectures and add Dropout at
rate $0.3$ and $0.5$. Note that the original implementation of VGG-16 has a
default Dropout at $0.5$, but here for the vanilla VGG-16 we remove the Dropout
for fair comparison.

\begin{figure*}
  \centering
  \includegraphics[width=.48\linewidth]{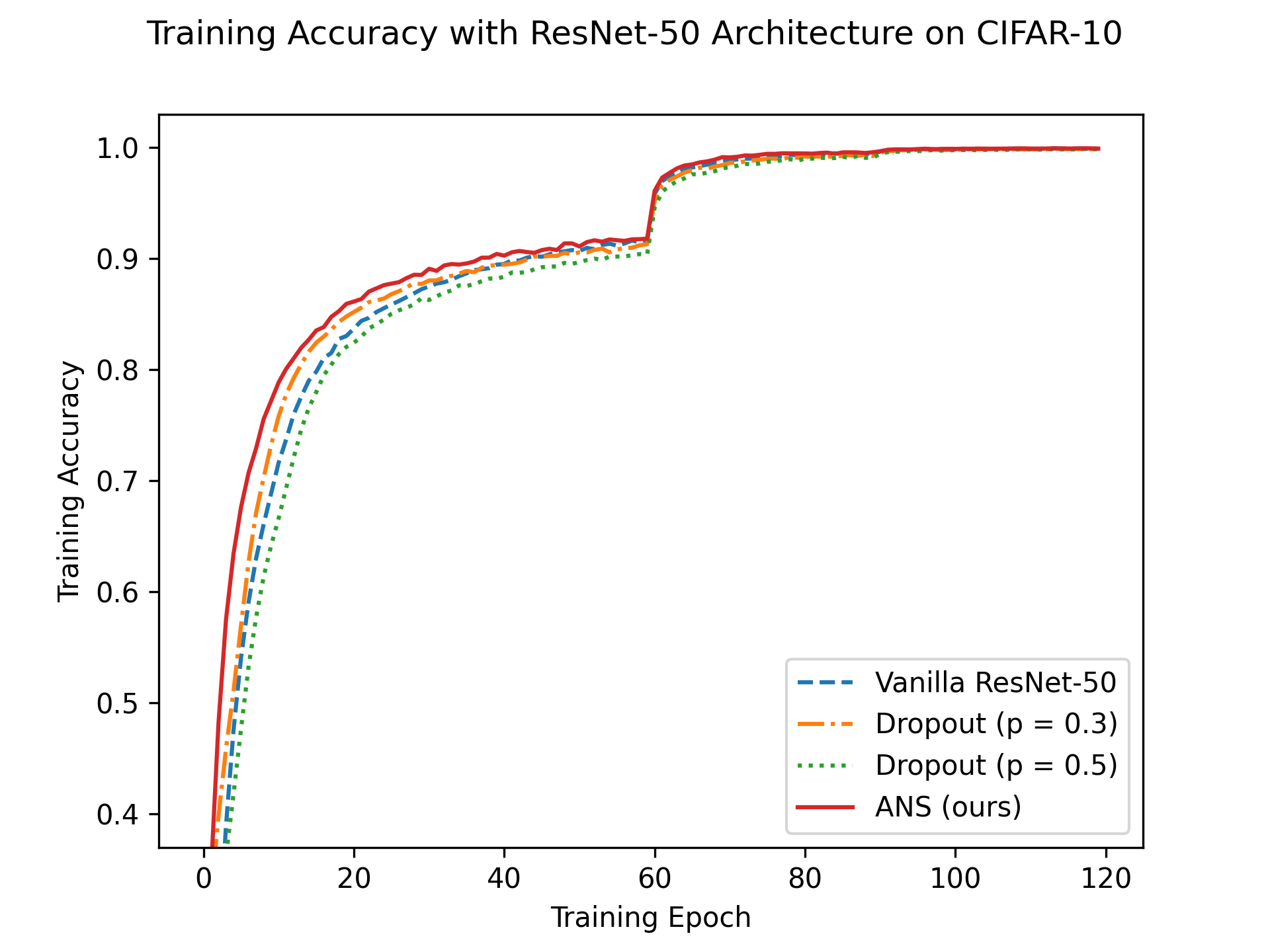}
  \includegraphics[width=.48\linewidth]{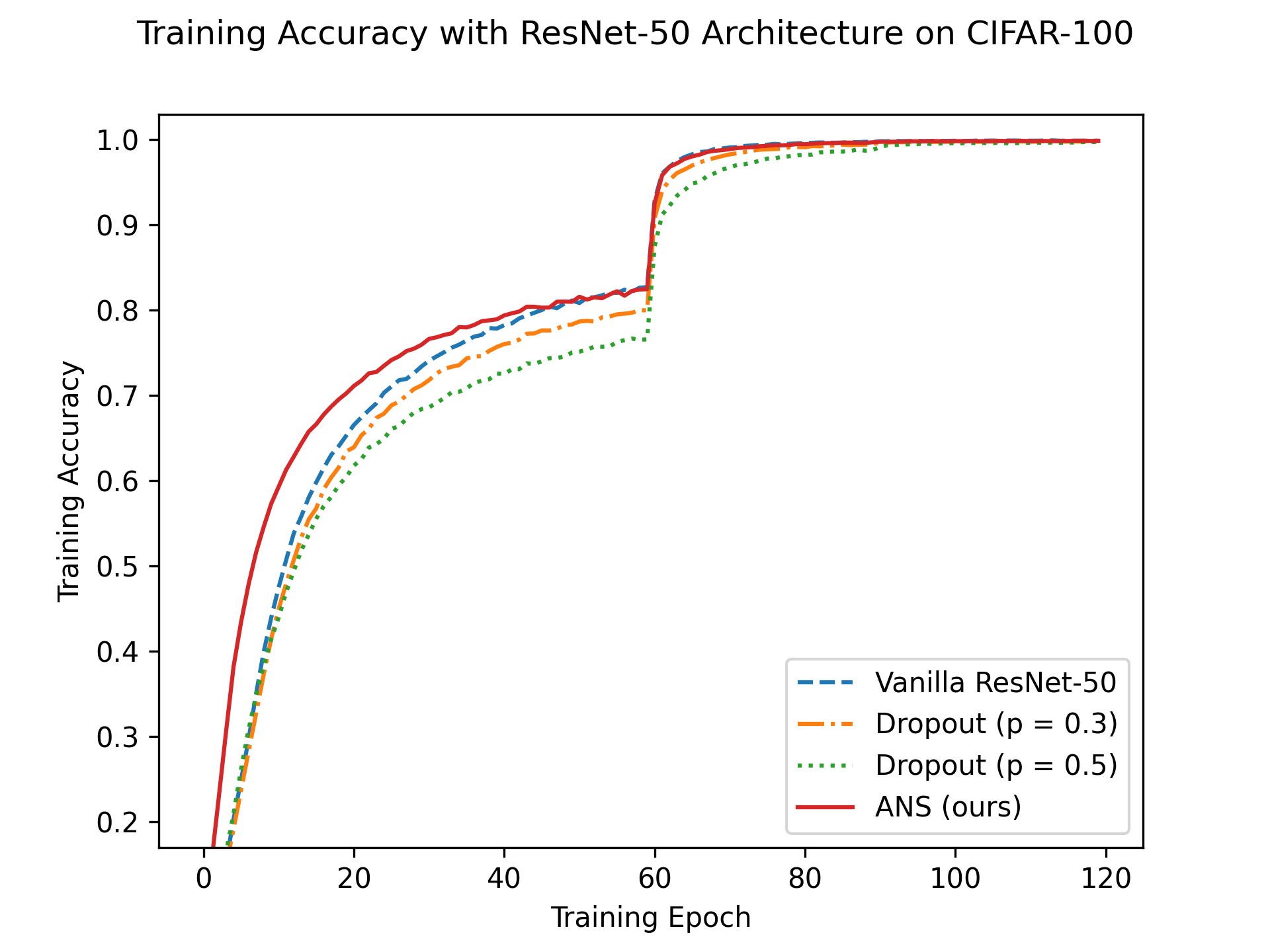}\\
  \vspace{2em}
  \includegraphics[width=.48\linewidth]{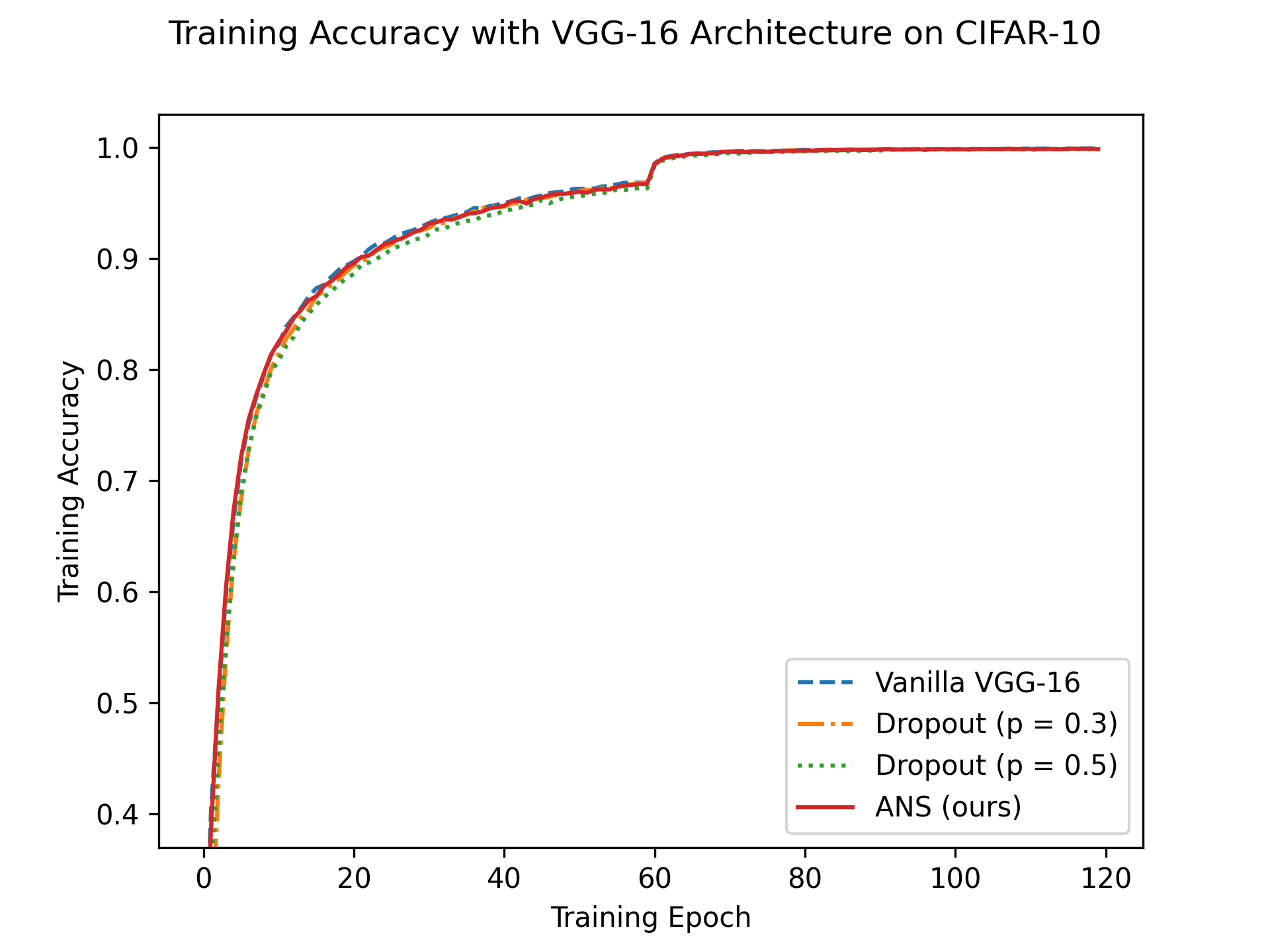}
  \includegraphics[width=.48\linewidth]{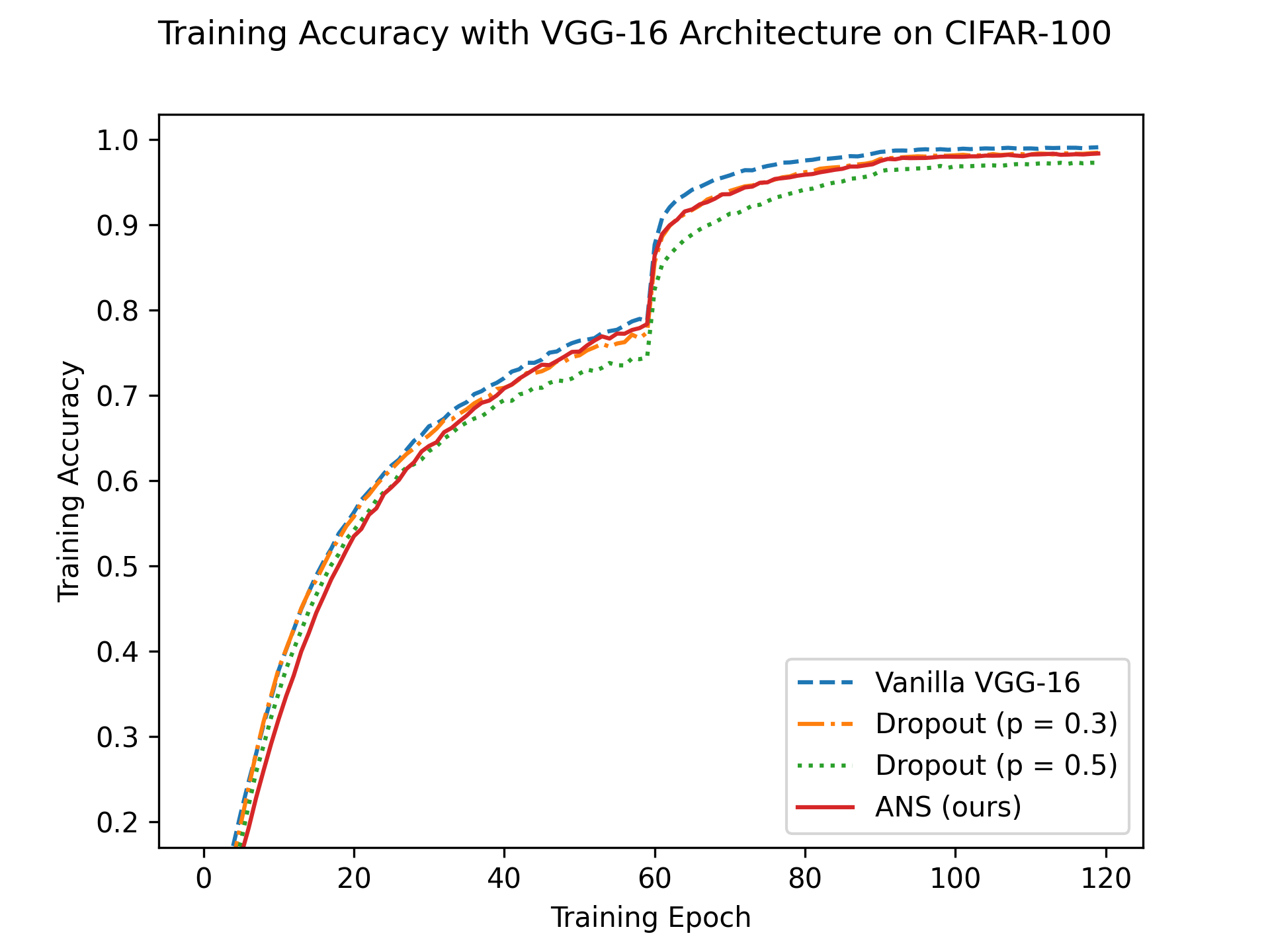}\\
  \caption{Visualization of training accuracy of the vanilla models, Dropout,
  and our method. While adding regularization to the model, our method achieves
  superior training performance on both datasets in comparison to Dropout. The
  training convergence is not slowed down and is sometimes even faster than the
  vanilla models. Better to be viewed in color.}
  \label{fig:train}
  \end{figure*}

\tabref{resnet} shows the test result with ResNet-50 on CIFAR-10 and CIFAR-100.
Since the ResNet-50 architecture does not have an intermediate fully connected
layer between the feature pooling layer and the classification layer, both
Dropout and ANS are applied only on the features layer after average pooling. We
can observe that Dropout does not show positive effect on the model performance.
However, our ANS method significantly improves the model performance, especially
on the harder benchmark CIFAR-100. 

For VGG-16, the results are shown in \tabref{vgg}. Given the VGG-16
architecture, we apply the Dropout and ANS on the last two fully connected
layers before the classification layer. The implementation with Dropout rate at
0.5 is identical to the original VGG-16 model. We can first observe that Dropout
significantly improves the VGG-16 model, because VGG-16 has more parameters and
connection that may be easier to cause overfitting. The proposed ANS method
further improves the model performance, outperforming the Dropout method. 

One of the drawbacks of methods such as Dropout and its variants is they require
specific design and tuning to work with different architectures. For example,
\cite{zaremba2014recurrent} shows that dropping neurons everywhere is not as
good as only dropping the neurons in the vertical connections for a
multi-layered LSTM network~\cite{hochreiter1997long}. Similarly, our results
show that even for similar ConvNet architectures, the application of Dropout
does not always benefit the model. However, our method consistently improves the
neural network despite the underlying architectures and datasets.

\subsection{Training Analysis}

In \figref{train}, we visualize the training process of the experiments,
comparing our method to the vanilla model and Dropout. For ResNet-50, we can
observe that with Dropout, the model has similar or slower convergence than the
vanilla model. However, our method achieves superior convergence. Since for
ResNet-50 the ANS method is directly applied on the final feature representation
layer after the pooling operation, it is able to improve the learning process by
putting more effort on the difficult cases. This is because the selection
pressure limits the usage of neurons for easier cases, and thus the loss on
difficult cases has larger weight on the network updates.

For the VGG-16 model, while we can still observe that the convergence of ANS is
slightly better than Dropout, the vanilla model actually converges faster.
Considering the fact that the vanilla VGG-16 has significantly worse testing
performance than using Dropout and ANS (as shown in \tabref{vgg}), we can see
that the vanilla VGG-16 easily overfits the training data if no regularization
method is used. This is because VGG-16 has many more parameters within the last
two fully connected layers. Regarding the complexity of the VGG-16 model, while
the convergence speed is similar, ANS improves the generalization capability of
VGG-16 with little cost on the training process.

\begin{figure*}
  \centering
  \includegraphics[width=.46\linewidth]{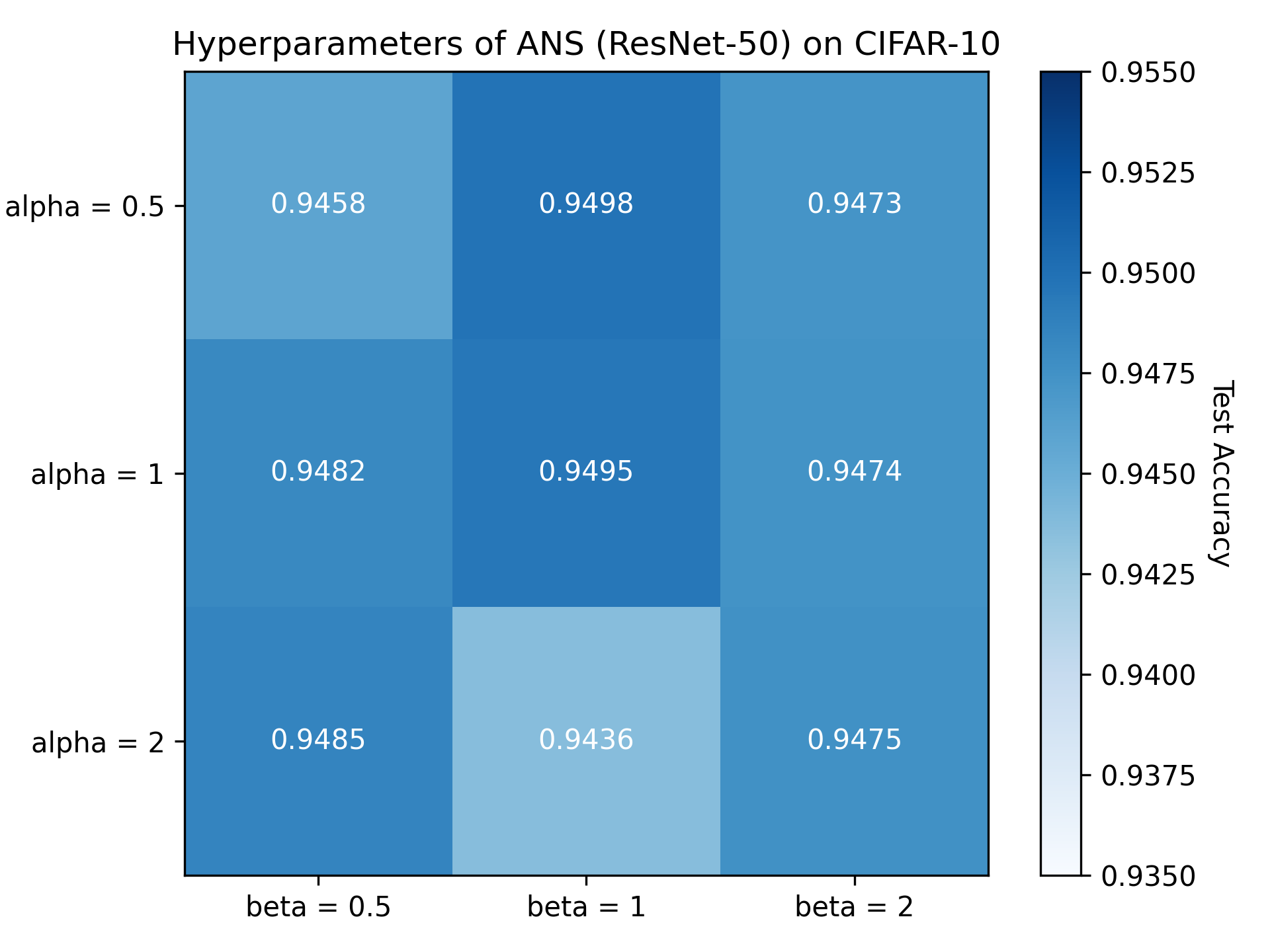}
  \includegraphics[width=.46\linewidth]{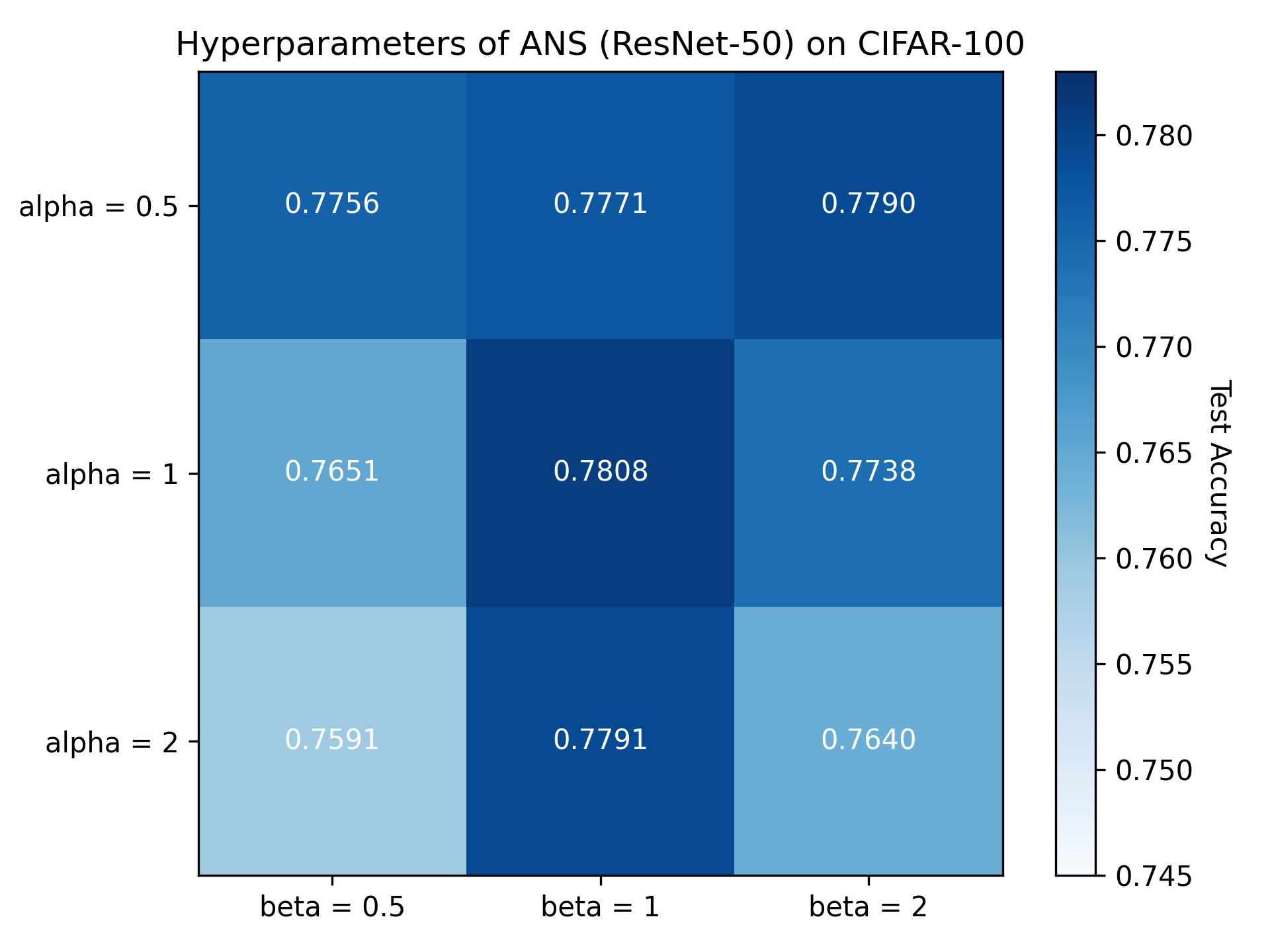}\\
  \vspace{2em}
  \includegraphics[width=.46\linewidth]{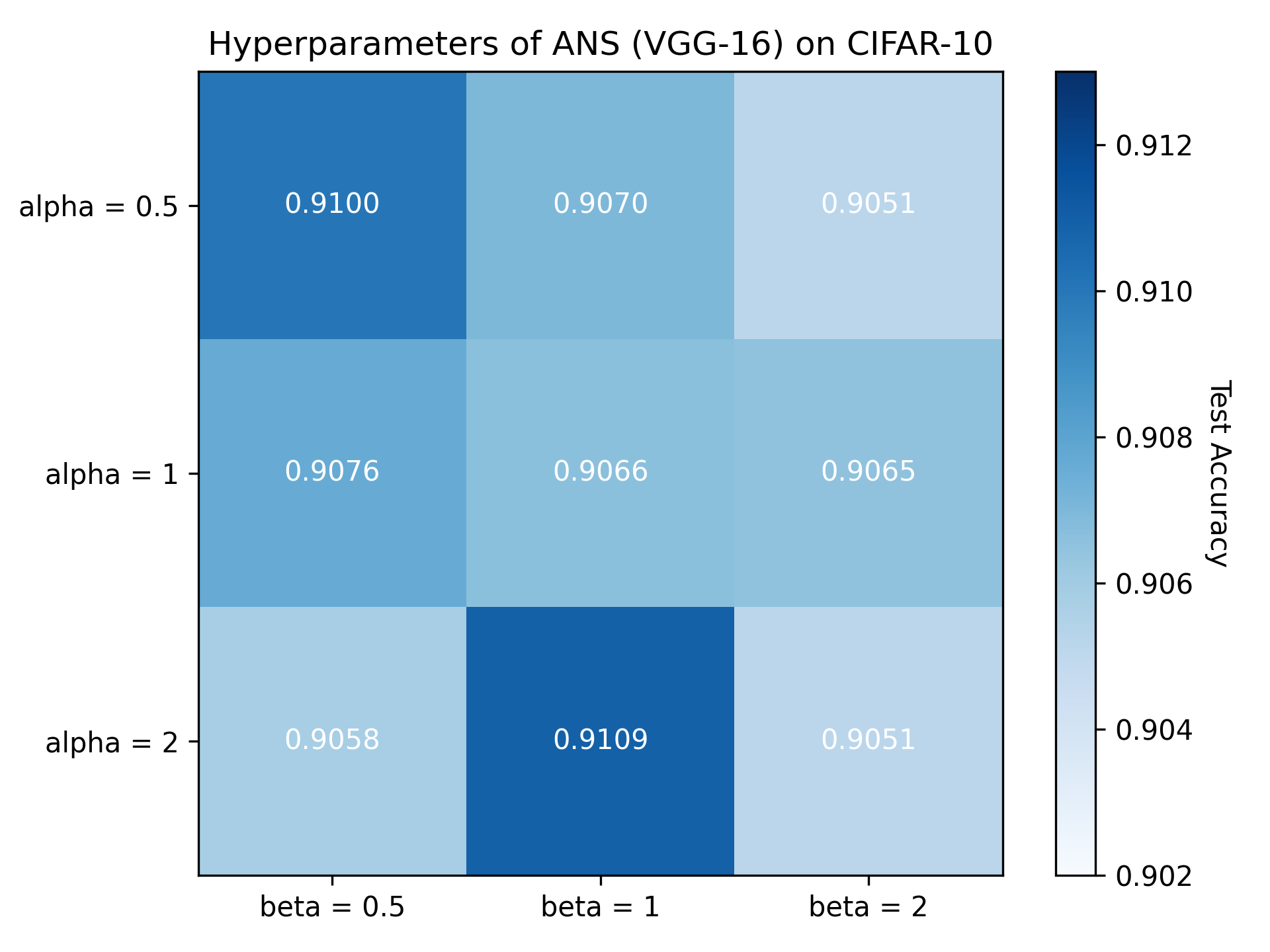}
  \includegraphics[width=.46\linewidth]{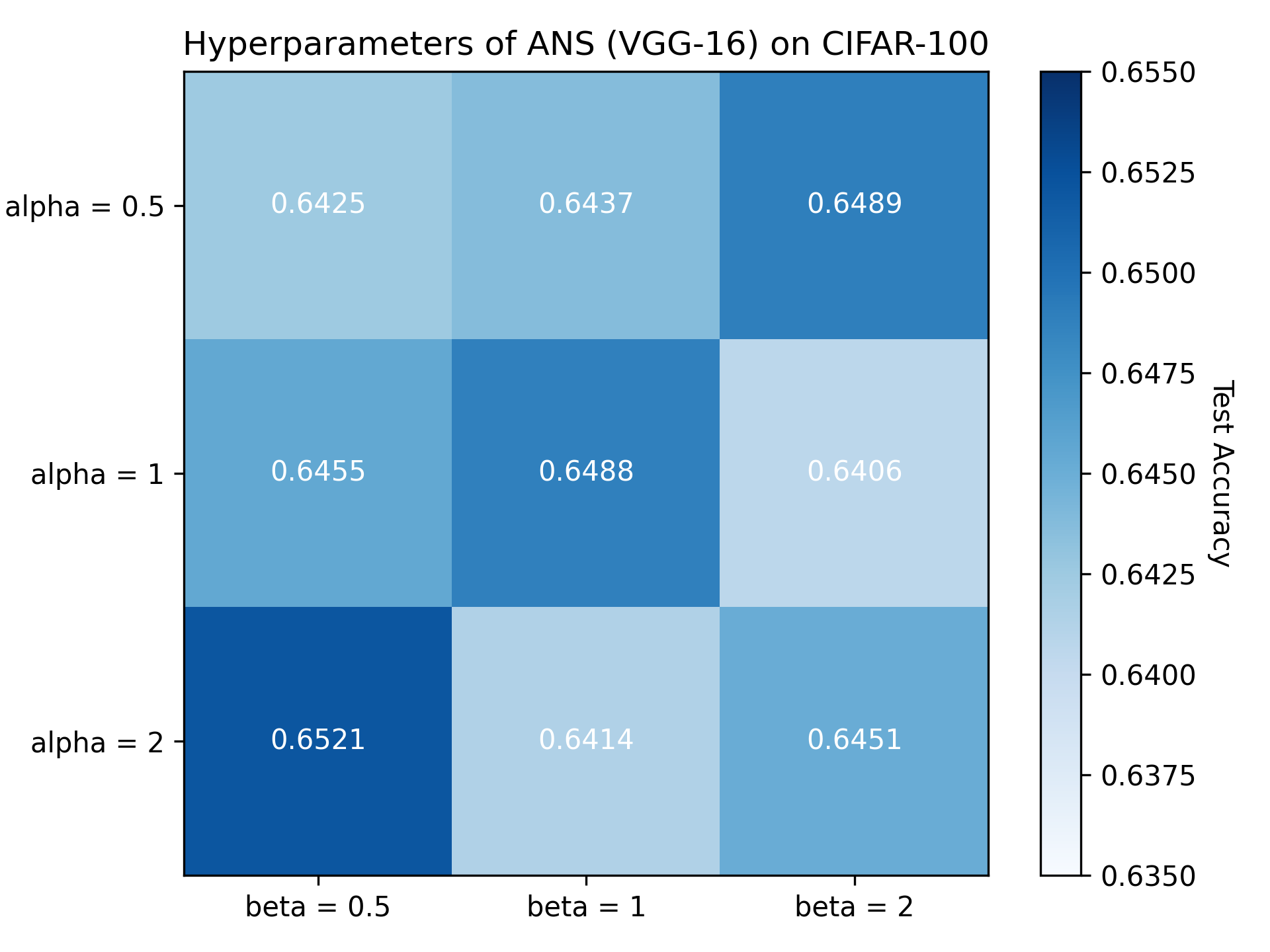}\\
  \caption{Classification results of ANS with different hyperparameter settings with  CIFAR-10 and CIFAR-100.}
  \label{fig:ab}
\end{figure*}

\begin{table}
  \caption{Classification results for ablation study with ResNet-50 on the test
   sets of CIFAR-10 and CIFAR-100.}
  \label{tab:ab_resnet}
  \begin{tabular}{lrr}
    \toprule
    Method & CIFAR-10 & CIFAR-100\\
    \midrule
    Vanilla ResNet-50  &  0.9418  &   0.7665\\
    ANS (self-attention only) & 0.9467  &   0.7700 \\
    ANS (full)     &  \textbf{0.9498}  &   \textbf{0.7808}\\
  \bottomrule
\end{tabular}
\end{table}

\begin{table}
  \caption{Classification results for ablation study with VGG-16 on the test
  sets of CIFAR-10 and CIFAR-100.}
  \label{tab:ab_vgg}
  \begin{tabular}{lrr}
    \toprule
    Method & CIFAR-10 & CIFAR-100\\
    \midrule
    Vanilla VGG-16     &  0.9046  &   0.6230\\
    ANS (self-attention only)  & 0.9088   &  0.6360\\
    ANS (full)         &  \textbf{0.9109}  &   \textbf{0.6521}\\
  \bottomrule
\end{tabular}
\end{table}

In addition, since CIFAR-10 has fewer classes and more data for each class, it is
usually easier to work with than CIFAR-100. Most of the analysis and results are
better illustrated on CIFAR-100.

\subsection{Ablation Study}

To further validate the effectiveness of the proposed ANS method, we perform
ablation study on the components and hyperparameters of ANS. The experiment settings are identical to the experiments in previous sections, where we test ANS with ResNet-50 and VGG-16 on both CIFAR-10 and CIFAR-100.

\subsubsection{Effectiveness of Components} The proposed ANS framework consists
of two components: the self-attention module and the adaptive regularization
loss. In this ablation study, we validate the effectiveness of each components
by having the test setting that removes the adaptive regularization loss by
letting $\alpha=0$, meaning the self-attention module is trained only using
gradients from the classification loss, without the adaptive regularization. The
results are shown in \tabref{ab_resnet} and \tabref{ab_vgg}.

We can first observe that with the self-attention module only, ANS is able to
improve the model performance by an obvious margin. Similar results have been
shown in some recent works~\cite{hu2018squeeze, vaswani2017attention} that
utilize attention-like mechanism in tuning deep neural network architectures.

Secondly, by adding the adaptive regularization, the performance is further
improved to a larger extent. For example, the testing accuracy is boosted by
$1.08\%$ by adding adaptive regularization to the self-attention-only ResNet-50,
while the self-attention itself adds $0.35\%$ to the vanilla model. The results
shows that the adaptive regularization mechanism is able to evolve better neural
selections comparing to just using end-to-end gradient training.

\subsubsection{Hyperparameters}

The fitness function of neural selection $\gamma$ is controlled by two
hyperparameters: $\alpha$ and $\beta$, as stated in Eq.~\ref{eq:gamma}. We
investigate how different hyperparameter settings may influence the model
performance with ANS, as the selection pressure is tuned by changing them.
\figref{ab} shows the testing results on different $\alpha$ and $\beta$ settings
in the format of heatmaps with the exact accuracy number in each cell. In order
to properly weigh the two losses for joint learning, we limit the scale of
searching space of $\alpha$ and $\beta$ by having $\alpha\in \{0.5,1,2\}$ and
$\beta\in\{0.5,1,2\}$. We test for every combination of the two hyperparameters
within this space.

In general, we can see that the framework is relatively robust to different
hyperparameters settings. For example, for the ResNet-50 model on CIFAR-100, the
worst-performing hyperparameter setting ($\alpha=2, \beta=0.5$) has testing
accuracy $0.7591$, which is still higher than using Dropout($p=0.3$) which has
accuracy $0.7587$. 

While there is no obvious pattern and correlation in terms of the
hyperparameters, it should be noted that the hyperparameters of ANS can be
problem dependent, just like hyperparameters for other selection methods in
evolutionary computing (such as the tournament size parameter for tournament
selection). However, we found that a setting of $\alpha=1$ and $\beta=1$ has
been robust for most of the cases.

\section{Conclusion and Future Work}

In this paper, we propose the Adaptive Neural Selection (ANS) method and a
framework for deep neural networks to evolve the behavior of selecting specific
neurons from the networks to perform the prediction task adaptively based on the
current input data. ANS can significantly improve the model performance on
generalization to unseen test cases without slowing down the training process.
Ablation study also shows that ANS has robust performance over different
hyperparameter settings.

While this work only applies ANS on fully-connected layers, ANS can be easily
extended to other network architectures, such as convolutional layers, by
transforming it to variants with specific structures. ANS can also be optimized
by using non-gradient neuroevolutionary techniques, which may further improve
the evolution of its weights.

In addition, while we keep the basic network architecture unchanged and only
evolve the selection of neurons, the ANS framework can also be used to evolve
the network architectures, serving as a strategy for neural architecture search.
For example, instead of putting regularization on attention weights, we can
alternatively use ANS to regularize the complexity of the network architecture
by removing layers, connections, \etc{} We look forward to a deeper analysis on
the effectiveness of combining ANS with other neural network evolution
strategies.

\begin{acks}

This material is based upon work supported by the National Science Foundation
under Grant No. 1617087. Any opinions, findings, and conclusions or
recommendations expressed in this publication are those of the authors and do
not necessarily reflect the views of the National Science Foundation. This work
was performed in part using high performance computing equipment obtained under
a grant from the Collaborative R\&D Fund managed by the Massachusetts Technology
Collaborative. The authors would like to thank Edward Pantridge, Anil Kumar
Saini, and Dr. Thomas Helmuth for their valuable comments and helpful
suggestions. 

\end{acks}

\bibliographystyle{ACM-Reference-Format}
\bibliography{sample-bibliography} 

\end{document}